\documentclass{article}

\usepackage{PRIMEarxiv}

\usepackage[utf8]{inputenc} 
\usepackage[T1]{fontenc}    
\usepackage{hyperref}       
\usepackage{url}            
\usepackage{booktabs}       
\usepackage{amsfonts}       
\usepackage{nicefrac}       
\usepackage{microtype}      
\usepackage{lipsum}
\usepackage{fancyhdr}       
\usepackage{graphicx}       
\graphicspath{{media/}}     

\title{Development of a Modular and Submersible Soft Robotic Arm and Corresponding Learned Kinematics Models
\thanks{
Department of Electrical and Computer Engineering, University of Illinois at Urbana-Champaign, Urbana, IL 61801, USA,\\
Beckman Institute for Advanced Science and Technology, University of Illinois at Urbana-Champaign, Urbana, IL 61801, USA,\\
Department of Nuclear, Plasma, and Radiological Engineering, University of Illinois at Urbana-Champaign, Urbana, IL 61801, USA,\\
Department of Nuclear Engineering and Radiological Sciences, University of Michigan, Ann Arbor, MI 48109, USA
}
}

\author{
  W. David Null \\
  Department of Electrical and Computer Engineering \\
  University of Illinois Urbana-Champaign \\
  Urbana, IL\\
  \texttt{null2@illinois.edu} \\
  \And
  Y Z \\
  Department of Nuclear Engineering and Radiological Sciences \\
  University of Michigan \\
  Ann Arbor, MI\\
  \texttt{yzyz@umich.edu} \\
}

\begin{document}
\maketitle

\begin{abstract}
Many soft-body organisms found in nature flourish underwater. Similarly, soft robots are potentially well-suited for underwater environments partly because the problematic effects of gravity, friction, and harmonic oscillations are less severe underwater. However, it remains a challenge to design, fabricate, waterproof, model, and control underwater soft robotic systems. Furthermore, submersible robots usually do not have configurable components because of the need for sealed electronics and mechanical elements. This work presents the development of a modular and submersible soft robotic arm driven by hydraulic actuators which consists of mostly 3D printable parts which can be assembled or modified in a relatively short amount of time. Its modular design enables multiple shape configurations and easy swapping of soft actuators. As a first step to exploring machine learning control algorithms on this system, we also present preliminary forward and inverse kinematics models developed using deep neural networks.
\end{abstract}


\section{INTRODUCTION}
Submersible soft robots have interesting potential use cases which include exploring the deepest parts of the ocean~\cite{13LiMarianaTrench2021}, interacting with delicate sea creatures~\cite{20LuongCoralReefs2019, 5GallowayDeepReefs2016, PhillipsDextrousGlove2018}, studying the locomotion methods of underwater animals~\cite{6MarcheseEscapeFish2014, 7ShenMollusc2017, 8ShenSoftArtificialMuscle2020, 9ShintakeFishDEAs2018, 10FengBodyWaveGener2020, 11KriegMarineInvert2015,12GiorgioMultiModal2017}, maintaining and inspecting underwater equipment in the presence of radiation~\cite{YirmiRadiationSoftRobots2019}, and creating safer and more versatile minimally invasive surgery procedures~\cite{4RuncimanMISSoft2019}. However, it remains a challenge not only to fabricate and maintain, but also to model and control underwater soft robotic systems. With these challenges in mind, this work presents the development of a modular hydraulically-actuated underwater 2D soft robotic arm. In addition to the robot's design, we present some preliminary forward and inverse kinematics models learned using neural networks.

Our modular design addresses common challenges which arise when designing soft robotic arms. Primarily, we call each segment in the arm a module and allow for arbitrary connection and disconnection of modules with minimal hardware alterations. Furthermore, any actuator on any module can be easily replaced. This is a desirable quality for research as actuators tend to break, leak, or show fatigue after long experiment sessions. It also may be necessary to test different actuator materials and it is helpful if this process is as painless as possible. Additionally, the electronics, sensing, valves, and fluid networks are local to each module of the robot. This allows the robot to take on multiple modular configurations while only needing one interconnecting hydraulic line. Also, each module is electrically connected to the same power lines and communication bus. 
\begin{figure}[h]
    \centering
    \includegraphics[width=0.6\columnwidth]{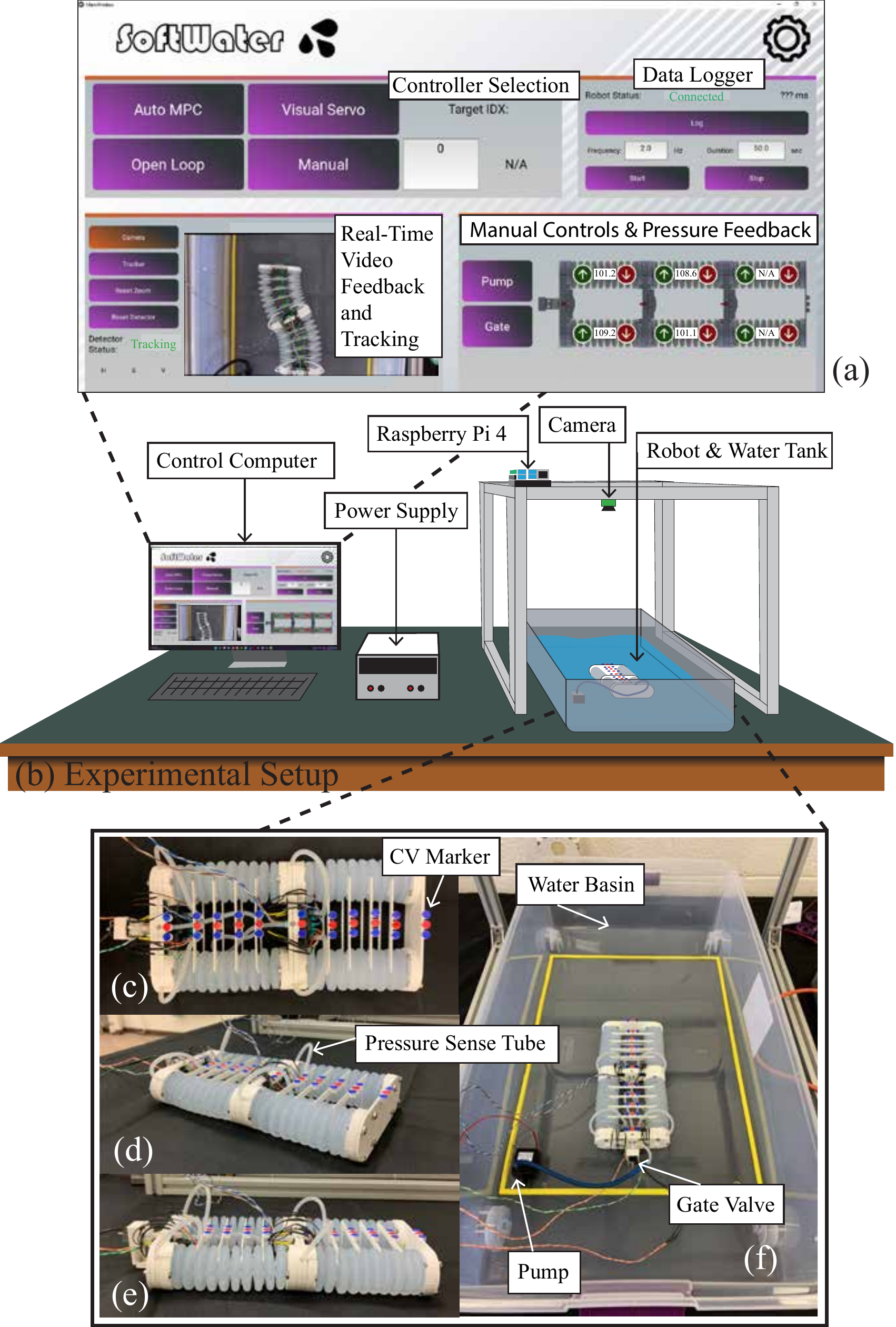}
    \caption{Desktop configuration of the underwater soft robot system. (a) Custom software for control, monitoring, and data logging. (b) Physical set up of the system on a workbench. (c) Top view of the robot with computer vision markers shown. (d) Isometric view of the robot with the pressure sensor tubes shown. (e) Side view of the robot. (f) Bird's eye view of the robot in the empty water basin with pump and gate valve shown.}
    \label{fig:exp_setup}
\end{figure}
\begin{figure*}[!t]
    \centering
    \includegraphics[width=\linewidth]{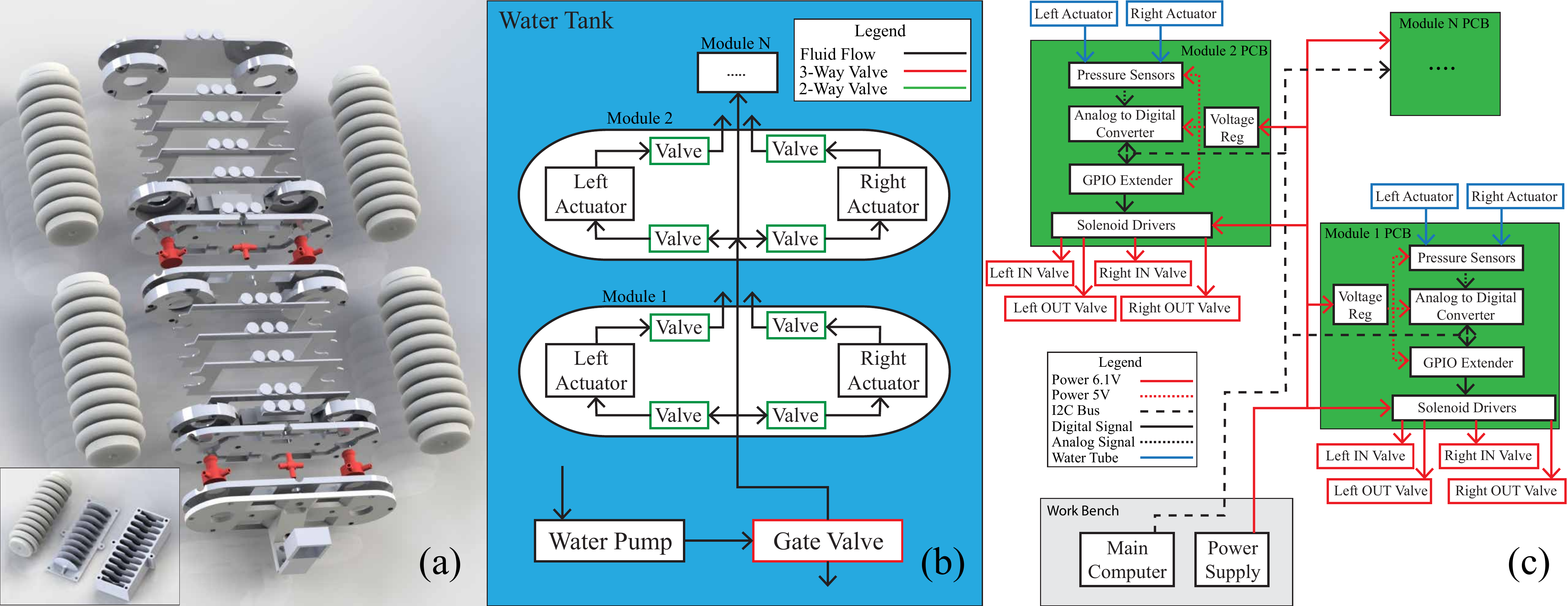}
    \caption{(a) Exploded view of the soft robot and render of the actuator and 3D printed mold used for fabrication. (b) Modular block diagram for the hydraulics system. (c) Modular block diagram for the electronics system.}
    \label{fig:modularity}
\end{figure*}

\begin{figure}[!t]
    \centering
    \includegraphics[width=\columnwidth]{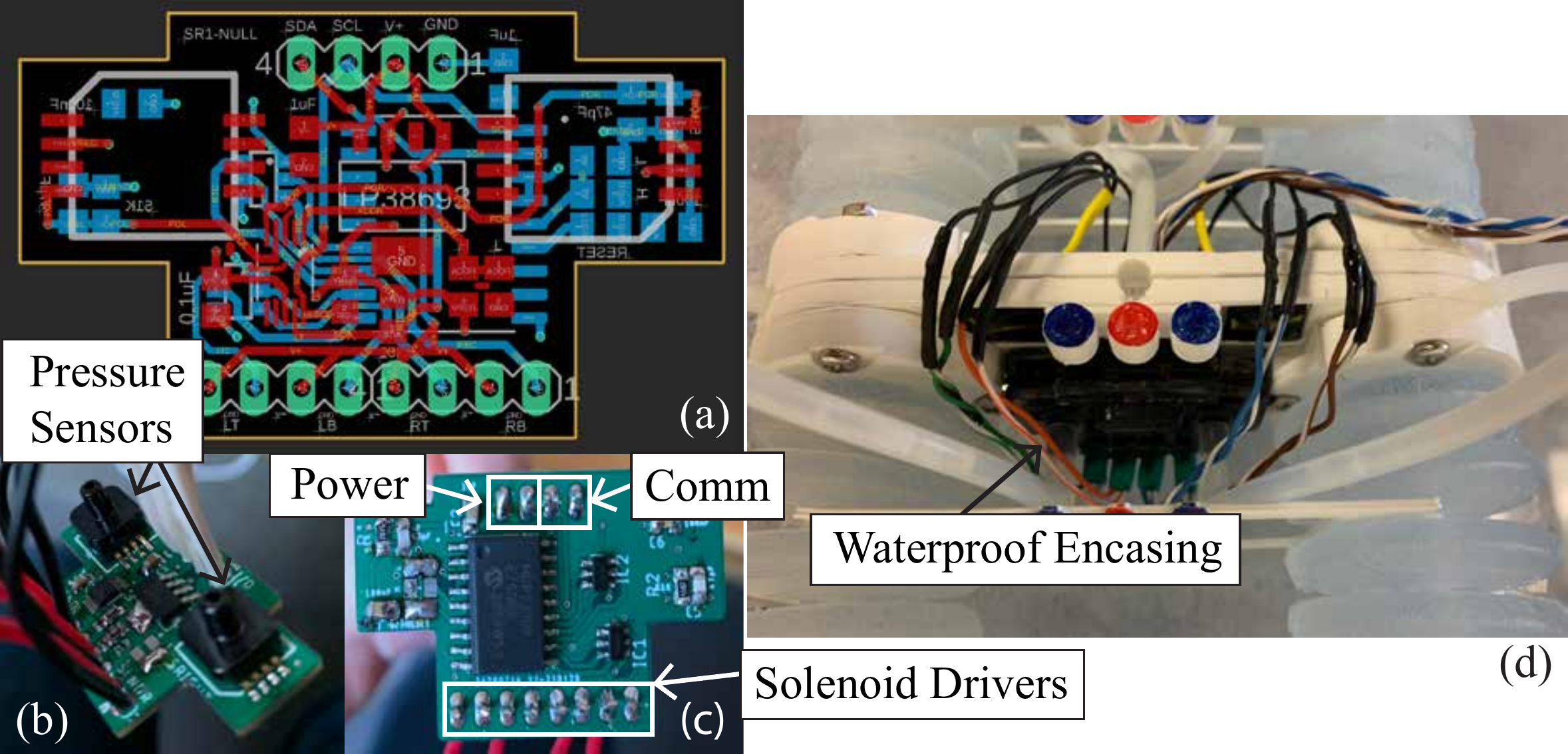}
    \caption{(a) Printed circuit board design. (b) Top of the board showing pressure sensors. (c) Bottom of the board showing solenoid driver outputs, power, and I2C communication connections. (d) Board in waterproof enclosure and installed on the robot.}
    \label{fig:pcb}
\end{figure}

To address the problem of modelling and controlling soft robots, promising methods are found in the literature using machine learning~\cite{ScheggModelingControlReview2022, 29ChinMLControl2020, 21BogueReviewUnderwaterRobots2015}. Neural networks can capture more nuances of nonlinear dynamics than analytical models such as piecewise constant curvature (PCC) and Cosserat rod theory. However, many learned models can also run in real time which is a challenge for models generated using finite element methods (FEM)~\cite{28WandkeMOOSE2021}. Inspired by the literature, this work presents preliminary kinematics modelling for the modular underwater soft robotic arm using neural networks. Both forward and inverse kinematics models are presented. These models form a benchmark for studying control techniques on this system, and in other work, the inverse kinematics model is used as a baseline controller in comparison to an auto-tuned data-driven model predictive controller~\cite{NullAutoMPCUnderwater2023}.  

\section{BACKGROUND}

Underwater environments are uniquely suited for the soft-body organisms we observe in nature and similarly for soft robots. This is partially due to the buoyant force of water counteracting the force of gravity and dampening the inherent oscillations that arise from actuating an elastic system. This work primarily focuses on soft hydraulic actuators~\cite{3BoyrazActuatorsSoft2018} which are fitting for remote mobile underwater robotics applications~\cite{19ShenUnderwaterBladders2020, 5GallowayDeepReefs2016, 16KurumayaWrist2018, 17SinatraUltragentle2019, 20LuongCoralReefs2019} because they only require a pump to drive actuation which can draw water from the surrounding environment. This gives them the advantage of continuous operation and also keeps the necessary external hardware light-weight and compact. Additionally, they have a relatively simple waterproofing process. In contrast, soft pneumatic actuators require pressurized air canisters with limited capacity, and cable-driven soft actuators typically require multiple electric motors each with precise sensing which need to be housed in a box external to the robot itself. 

Researchers have exploited these advantages by implementing several underwater soft robotic arms and manipulators~\cite{5GallowayDeepReefs2016, 15GongBioinspiredSoftArm2018, 16KurumayaWrist2018, 17SinatraUltragentle2019, 18GongIKMethon2018, 19ShenUnderwaterBladders2020, 20LuongCoralReefs2019}. Miniaturized versions of these robots have potential use cases in minimally invasive surgery (MIS)~\cite{4RuncimanMISSoft2019} where the robot can be surrounded by blood~\cite{22XuUnderwaterCableDriven2018}. Furthermore, insights on the characteristics of underwater soft robots can also be applied more generally as Du et al. showed by improving a simulation model developed for an underwater robot~\cite{23DuDiffSim2020}. 

Hydraulically actuated underwater soft robots have immense potential, however, designing accurate and robust controllers for these systems remains a challenge despite their natural advantages. High nonlinearity and large degrees of freedom render analytical models, such as piecewise constant curvature (PCC)~\cite{24WebsterPCCRev2010}, insufficient for many real-world applications. The finite element method has been applied successfully to simulate soft robots~\cite{27RungePCCFEM2017} but suffers from large computational complexity which limits its usability in real time~\cite{28WandkeMOOSE2021}. In traditional controller design, feedback techniques are used to compensate for modelling errors. However, despite exciting new developments in soft sensors research~\cite{25ThuruthelRNNsense2019, 26JungElastomerOptical2019}, the sensors themselves are not readily available, and are difficult to fabricate, install, and calibrate.  

At this point, researchers have turned to data-driven modelling and control strategies to capture the nuances of a system’s nonlinearity, while running in real time~\cite{29ChinMLControl2020}. A widely-used method is learning the forward and inverse kinematics of a soft robot with a neural network~\cite{30RolfElephantTrunk2014}. Deep neural networks and recursive neural networks have also been used to develop a dynamics model for use in a model predictive controller~\cite{31GillespieMPCSoftRobot2018, 32HyattMBCSoftActuators2019, 33ThuruthelOpenLoopSoftRobotic2017}. Additionally, reinforcement learning has been proposed as a model-free approach to the problem~\cite{29ChinMLControl2020, 34YouSoftRL2017}. An issue with these approaches is that a disturbance or change to the robot's weight can cause the estimated models or policies to break down for lack of sufficient data. An interesting approach to solving this problem using online learning with Gaussian Process regression has been proposed to account for disturbances in system dynamics~\cite{35FangVisionGPR2019}.

\section{MATERIALS AND METHODS}
This work presents the development of a submersible soft robotic arm with a modular design which is simple to assemble, and useful for research. The mechanical structure consists of 3D printed parts, and the actuator molds can also be 3D printed. The robot can collect internal state data from pressure sensors and solenoids positioned locally to each actuator and synchronize this with ground truth positioning information gathered from an overhead camera. The soft hydraulic actuators can be easily swapped with those of different shapes, sizes, and material characteristics. Additionally, the robot's overall structure consists of individual segments, which we call modules, whose electronics and fluid networks can attach to each other. These modules can be added and subtracted to reach a desired robot configuration. This work also presents the development of preliminary learned forward and inverse kinematic models for use in robot modelling and control. 

\subsection{Experimental Setup}
The entire system was built to work in a desktop configuration for rapid algorithm development and testing. In the desktop configuration, the robot's base is mounted to a sheet of acrylic and submerged in about 5 cm of water along with the water pump which creates the hydraulic pressure (Fig. \ref{fig:exp_setup}(f)). Computer vision (CV) markers are arrayed along the robot's centerline (Fig. \ref{fig:exp_setup}(c)) which are used to track the true shape of the robot. The water level is such that the CV markers rest just above the surface so they can easily be seen by the camera positioned directly above the robot (Fig. \ref{fig:exp_setup}(b)). Computer vision was chosen over motion capture because of the issue of infrared light reflecting off the surface of water, and the fact that cameras are more deployable in real-world applications. The camera and the robot are both controlled by a Raspberry Pi 4 computer for this project. 

The robot presented in this work consists of two stacked modules. Both modules consist of two soft hydraulic actuators, and each actuator is driven by two solenoid valves (in/out) and sensed by one pressure sensor. The actuators are fabricated out of a silicon polymer using 3D printed molds (Fig. \ref{fig:modularity}(a)). A pump creates water pressure which is channeled through a network of tubes and valves (Fig. \ref{fig:modularity}(b)). On each module's local circuit board, there are two analog pressure sensors. It is important to keep the distance between actuator and pressure sensor small (Fig. \ref{fig:exp_setup}(d)) to ensure an accurate and responsive reading. The camera is positioned directly above the robot (Fig. \ref{fig:exp_setup}(b)), and the samples from the pressure sensors are synchronized with photos taken from the camera. 

\subsection{Modular Design} 
One of the key features of this soft robotic system is its modular design. The electronic, mechanical, and hydraulic components are all built so that modules can be stacked on top of each other to form different configurations for different experiments and applications. Similarly, the actuators themselves can be swapped out easily. This enables easier maintenance as well as quicker iterations on shapes, sizes, and material properties for testing. The system's modularity creates potential for many different experimental parameters and applications.

\subsection{Mathematical Representation of Robot} 
\begin{figure}[!t]
    \centering
    \includegraphics[width=0.85\columnwidth]{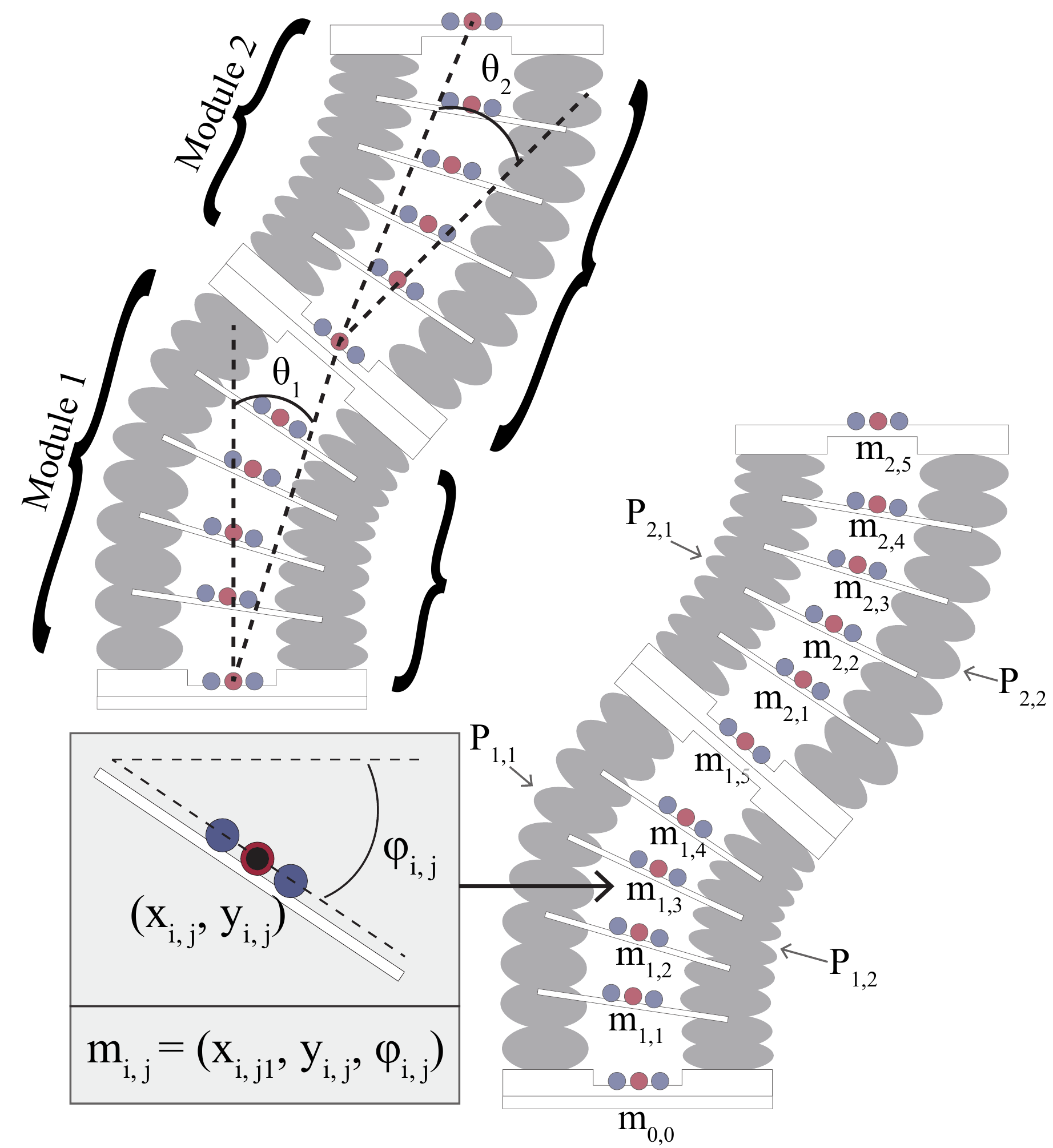}
    \caption{Mathematical representation of the robotic system showing modules, bend angles (top left), computer vision marker array, and pressure values (bottom right).}
    \label{fig:math_rep}
\end{figure}
A mathematical representation of the robot is presented in Fig. \ref{fig:math_rep}. The robot is constructed of two modules with the bottom-most marker of the base module ($m_{0,0}$) representing the origin. Each module has five CV markers labelled $m_{i,j}$ where $i\in \{1,2\}$ and $j \in \{1,2,3,4,5\}$. Each CV marker contains three circles whose centroid locations can be identified by the overhead camera. The red circle in the middle represents the center of that section of the robot and is given an $(x,y)$ location. All the red circles together represent the underlying centerline shape of the robot. The blue markers make it possible to receive the orientation of that section of the robot with respect to the horizontal. This orientation is represented by an angle $\phi$. All this information is captured in one marker tuple as $m_{i,j} = (x_{i,j}, y_{i,j}, \phi_{i,j})$. The pressure value for each actuator is denoted by the variable $P_{a,b}$ where $a \in \{1,2\}$ for modules $1$ and $2$ and $b \in \{1,2\}$ for left and right.

\subsection{Basic Actuation Experiment}
\begin{figure}[!t]
    \centering
    \includegraphics[width=\columnwidth]{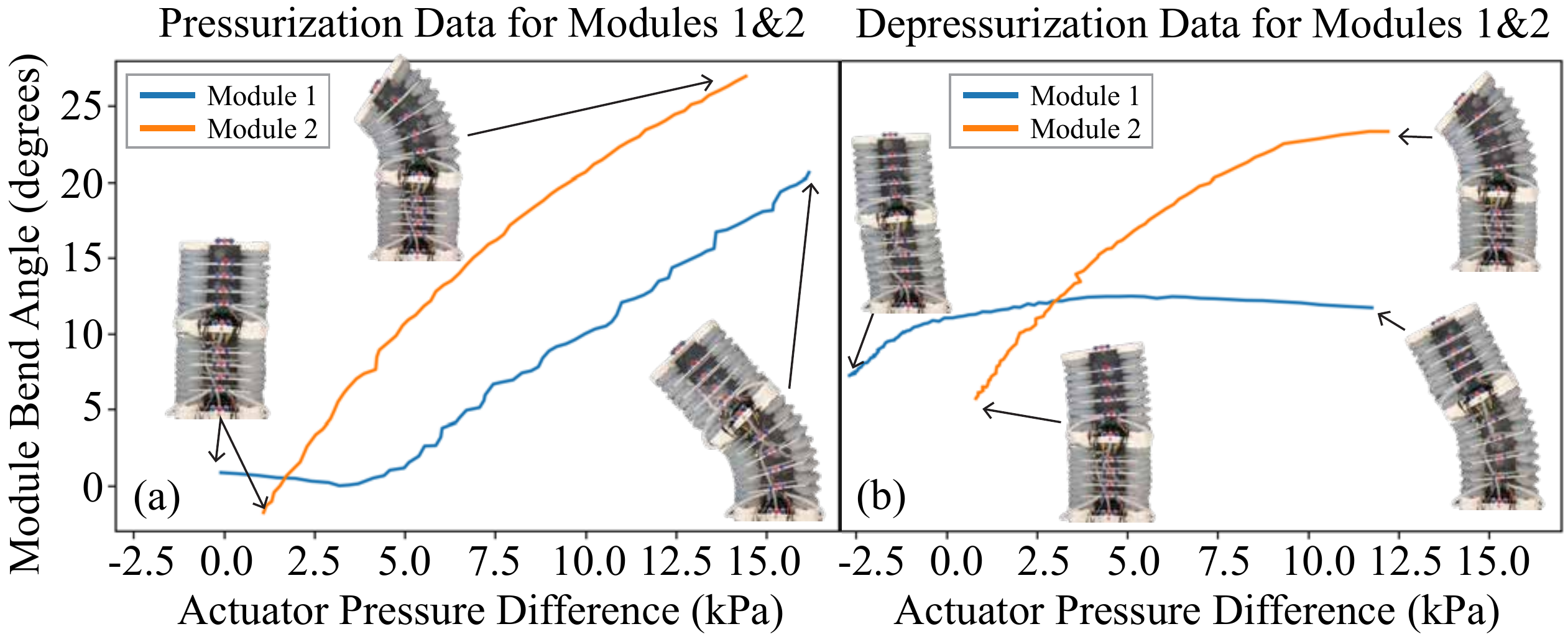}
    \caption{Plots showing the bend angles of modules 1 and 2 corresponding to $(\theta_1, \theta_2)$ as a function of the pressure difference in each of their actuators as the right actuator is pressurized for 31 seconds in (a), and depressurized for 42 seconds in (b). A hysteresis was observed in both modules.}
    \label{fig:basic_act}
\end{figure}

Experimental data shows the nonlinear characteristics and hysteresis associated with this robot. Fig. \ref{fig:basic_act} shows the bend angles $\theta_1, \theta_2$, defined in Fig \ref{fig:math_rep}, of each module as a function of the pressure difference between that module's actuators. The pressure difference is represented by the equation $P_{\textrm{diff}} = P_{i,2} - P_{i,1}$ where $i \in \{1,2\}$ for modules 1 and 2. Depressurization relies on the elasticity of the actuators to force water out and therefore is a longer process than pressurization. In Fig. \ref{fig:basic_act}, the pressurization data is taken over the course of 31 seconds vs. 42 seconds for the depressurization data. In both experiments, only the right actuator was pressurized or depressurized. 

These plots give us insights into the dynamics of the robot. In a similar way to the force of a spring, as the actuator shrinks the force causing it to shrink decreases. This leads to the different ending shapes for the pressurization and depressurization curves. Secondly, module 1 demonstrates more severe hysteresis that module 2. This is shown by the fact that the angular displacement for module 1 $(\theta_1)$ does not start increasing during pressurization until the pressure difference is nearly $3.7$ kPa. Whereas the angular displacement starts increasing for module 2 $(\theta_2)$ almost immediately when the pressure difference starts increasing. A similar trend is also seen in the depressurization curves. This is explainable because module 1 needs to drag along module 2 wherever it goes and therefore requires more force to start moving. Finally, it is interesting that the pressure difference goes negative at the end of depressurization for module 1 shown in (Fig. \ref{fig:basic_act}(b)). This means that the pressure for the left actuator is greater than the pressure in the right actuator by 2.5 kPa even though no water entered the left actuator. This happens because as the right actuator shrinks, it compresses the left actuator, increasing its pressure because its out-valve is closed. All these system characteristics are difficult to model analytically with sufficient accuracy. However, it is possible to learn these dynamics as shown in the next sections.

\subsection{Preliminary Learned Kinematics Models}
Using the data collected, two deep neural networks (DNNs) were designed, trained, and tested to model the forward and inverse kinematics of the submersible soft robot. The forward kinematics model takes in previous and current robot state parameters and outputs estimated CV marker positions for the centerline markers. The inverse kinematics model takes as inputs the centerline marker positions for a single time step and outputs estimated pressure values for each actuator. 

Both networks were built in Python using the Keras package. Having a similar internal structure, they consist of 4 dense hidden layers each of which uses a relu activation function and is followed by a dropout layer with a probability value of 0.2. The dense layers have a size of 128, 64, 32, and 16 neurons respectively. The difference between the forward and the backward model are the input and output dimensions. In total, the forward network has 17,476 trainable parameters while the backward network has 13,620 trainable parameters. Both networks are trained with a mean squared error loss function and the ``adam'' optimizer.
\subsubsection{Forward Kinematics Model}
The shape of the robot at time $t$ is highly dependent on previous and current measurements of pressure in each actuator and the states of the in/out solenoid valves. We represent the actuator pressures at time $t$ by the vector $\mathbf{p}_t \in \mathbb{R}^i$ where $i=4$ for the 2-module robot configuration. Similarly, the binary states of the in/out solenoid valves at time $t$ are contained in the vector $\mathbf{u}_t \in \mathbb{R}^j$ where $j=8$ for the 2-module configuration. To achieve higher accuracy, we also consider previous time-steps with a back-step size of $\tau=7$ and a total number of previous samples $n=3$. Since data was collected at a frequency of $2$ Hz, the DNN will take as input $4$ samples of $\mathbf{p}$ and $\mathbf{u}$ over the last $(7)(3)(0.5)=10.5$ seconds. The output of this neural network are the estimated $x$ and $y$ coordinates at time $t$ of the centerline markers. Thus, the learned forward kinematics model is represented by $\hat{f}_{FK}$ and is used as follows:
\begin{equation}
\hat{\mathbf{m}}_{t} = \hat f_{FK}( \mathbf{p}_{t}, \mathbf{p}_{t-\tau}, ..., \mathbf{p}_{t - n\tau}, \mathbf{u}_{t},\mathbf{u}_{t-\tau}, ..., \mathbf{u}_{t-n\tau})
\end{equation}
\begin{figure}[h]
    \centering
    \includegraphics[width=\columnwidth]{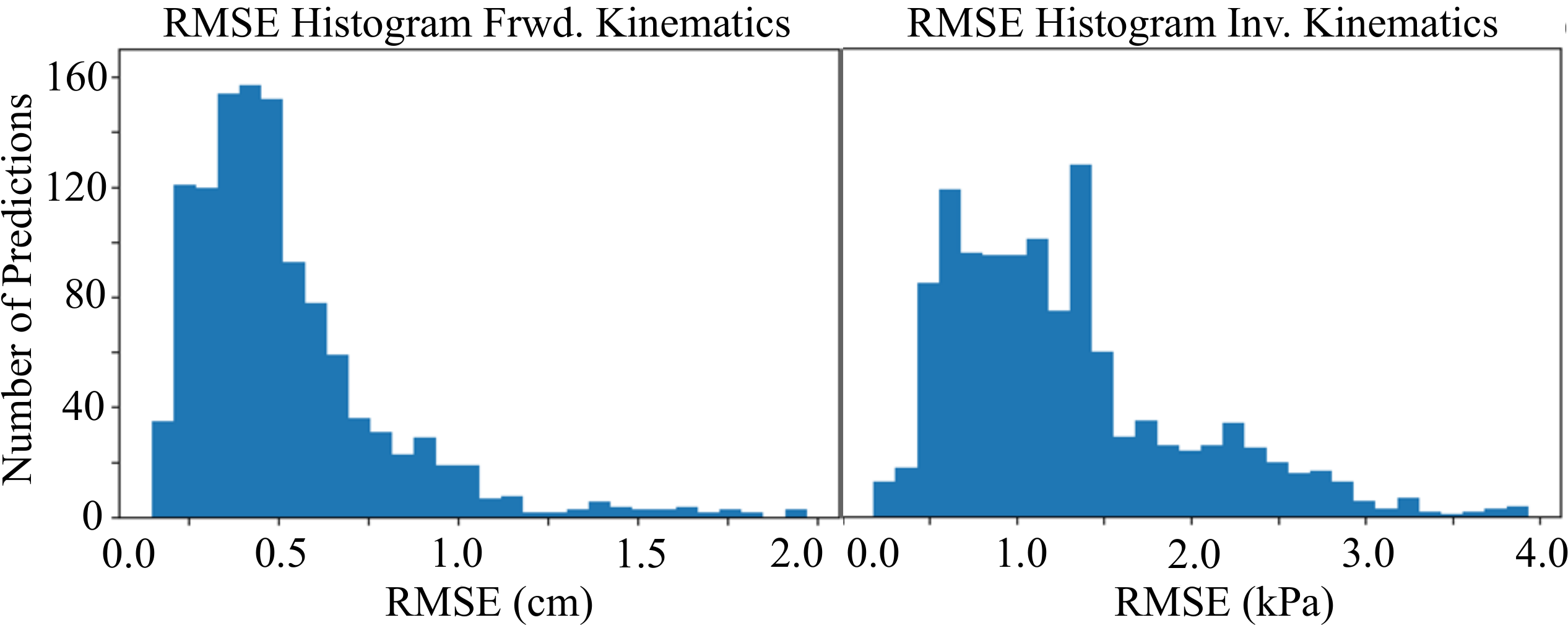}
    \caption{Histogram of the root mean squared error (RMSE) over the validation data set for the forward kinematics model (left) and the inverse kinematics model (right).}
    \label{fig:hist}
\end{figure}
\begin{figure*}[!t]
    \centering
    \includegraphics[width=\linewidth]{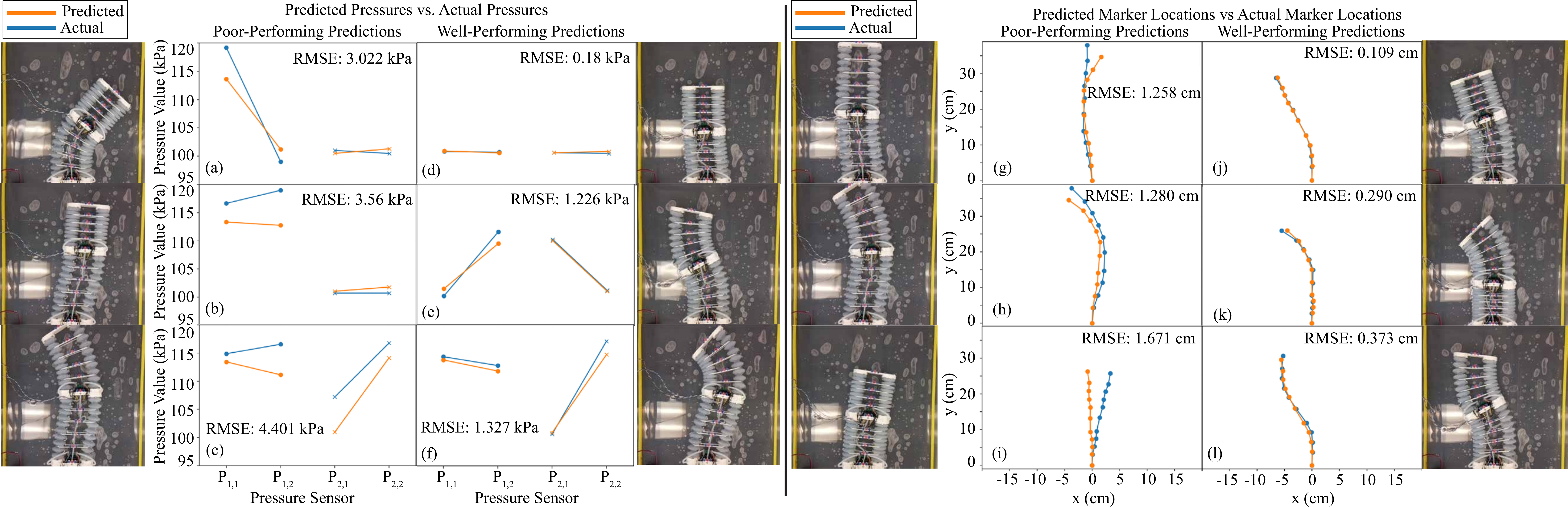}
    \caption{Poor-performing (a-c) and well-performing (d-f) pressure estimations generated by the learned inverse kinematics model. Poor-performing (g-i) and well-performing (j-l) estimated shape configurations generated by learned forward kinematics model.}
    \label{fig:pred}
\end{figure*}

\subsubsection{Inverse Kinematics}
In keeping with how inverse kinematics models are typically used in practice, our learned model simply takes as inputs the centerline marker $x$ and $y$ coordinates corresponding to a desired shape represented by $\mathbf{m}_t \in \mathbb{R}^{k\times 2}$ where $k=10$. The model outputs an estimated pressure vector $\hat{\mathbf{p}}_t \in \mathbb{R}^i$ where $i=4$. Since this model does not consider previous state data, it will be less accurate, but more useful in practice because the programmer only needs to specify a desired shape and not a full shape trajectory. Thus, the learned inverse kinematics model is represented by $\hat{f}_{IK}$ and is used as follows:
\begin{equation}
\hat{\mathbf{p}}_{t} = \hat f_{IK}( \mathbf{m}_{t})
\end{equation}

\subsubsection{Data Collection}
To generate data for training the models, the robot was manually actuated along multiple preplanned trajectories which brought each actuator to its pressure limit in different configurations with other actuators. Each run lasted between 8-17 minutes and was sampled at 10 Hz. In total, about 50 minutes of data was collected. All data was normalized in preparation for training. All pressure values were normalized between 95 kPa and 121 kPa, while all marker x-coordinates were normalized between -15 cm and 15 cm, and all marker y-coordinates were normalized between 0 cm and 40 cm. 

\section{RESULTS AND DISCUSSION}
\subsection{Model Training and Performance}
The 5,886 collected data points were shuffled and split 80\%/20\% for training and testing respectively. Both models were trained over 100 epochs with a batch size of 8. Fig. \ref{fig:hist} shows the histograms of the root mean squared errors (RMSE) when evaluating both models on the same testing data points. The forward model performs very well with about $60$\% of the shape estimations averaging less than $5$ mm of error per CV marker position. The errors for the inverse kinematics model are mostly concentrated below $1.5$ kPA of error.

\subsection{Discussion}
Examples of well-performing and poor-performing estimates given be the learned forward kinematics model are shown in Fig. \ref{fig:pred}. The poorest estimates of the learned forward kinematics model only reach about $~1.5$ cm RMSE. These poor-performing cases seem to occur at the limits of the robot's task space as in Fig. \ref{fig:pred}(g,h), or near the fully depressurized state as in Fig. \ref{fig:pred}(i). In the cases where the robot is near the limits of its task space, the issue could lie in not having enough training data for those regions.  In the fully depressurized instance, the actuators in the first module are not able to drag the second module all the way back to the centered home position. Thus, the pressure is equalized for an extended period, but the position of the robot is more uncertain. A potential solution to this issue would be to increase the stiffness of the actuators on the first module.

While the learned inverse kinematics model was not as accurate as the forward kinematics model, it was also not given any previous states as inputs. This is because it is not always possible to track the shape of the robot over time. Often the robot arm will be operating in a space where a camera cannot see its full shape. Thus, only the target shape and the robot's current pressure sensor and solenoid valve states are known. Even with simple inputs, the model performs relatively well. Examples of poor-performing and well-performing estimations is shown in Fig. \ref{fig:pred}. A variation of this model is used in a benchmark controller for the robot in \cite{NullAutoMPCUnderwater2023}. 

\section{CONCLUSION}
This work presents an underwater soft robotic arm that is not only relatively simple to assemble but also modular and configurable and can be used in a desktop environment. The electronics and fluid networks in the robot are designed so that segments can be added or removed to form a desired robot size. Also, the mounting method used to attach the soft actuators allows for swapping of different types of actuators which may vary in size, shape, and material characteristics. This also allows an actuator to be easily maintained or replaced if it leaks or degrades over time. In addition, this robot arm can also be used to develop machine learning models for itself. 

Some of the robot's current limitations motivate upgrades to the design as future work such as experimenting with different size pumps, and different actuator shapes, sizes, and stiffnesses. Adding a 3rd actuator to each module could enable 3D motion. Installing the arm on underwater ROV and using it at different depths could lead to interesting nuances in the dynamics. Also, miniaturizing the design may lead to research applications related to minimally invasive surgery. 

There are also exciting research directions that can build on the learned models presented in this work. Reinforcement learning control algorithms could be developed using the learned forward kinematics model to generate policy training data. These algorithms could be compared with data-driven model predictive controllers as well. Finally, a gripper could be attached to the end of the robot and various online learning strategies could be investigated to compensate for changes in dynamics when the robot grasps an object. 

\bibliographystyle{unsrt}  
\bibliography{references}

\end{document}